\pdfoutput=1
\documentclass[11pt]{article}

\usepackage{acl}

\usepackage{times}
\usepackage{latexsym}
\usepackage{booktabs}
\usepackage{bbm}
\usepackage{multirow}

\usepackage[T1]{fontenc}
\usepackage{amsmath}
\usepackage{amssymb}
\usepackage[utf8]{inputenc}

\usepackage{microtype}

\usepackage{inconsolata}

\usepackage{graphicx}

\title{Augmenting Document-level Relation Extraction with Efficient Multi-Supervision}

\author{Xiangyu Lin$^1$, Weijia Jia$^{2}$ and Zhiguo Gong$^1$$^*$ \\
         $^1$SKL-IOTSC, University of Macau$^\dagger$\\  
         $^2$BNU-UIC \& Beijing Normal University (Zhuhai)\\
         \small\texttt{yc07403@um.edu.mo,  jiawj@uic.edu.cn, fstzgg@um.edu.mo}}

\begin{document}
\maketitle
\begin{abstract}
Despite its popularity in sentence-level relation extraction, distantly supervised data is rarely utilized by existing work in document-level relation extraction due to its noisy nature and low information density. Among its current applications, distantly supervised data is mostly used as a whole for pertaining, which is of low time efficiency. To fill in the gap of efficient and robust utilization of distantly supervised training data, we propose Efficient Multi-Supervision for document-level relation extraction, in which we first select a subset of informative documents from the massive dataset by combining distant supervision with expert supervision, then train the model with Multi-Supervision Ranking Loss that integrates the knowledge from multiple sources of supervision to alleviate the effects of noise. The experiments demonstrate the effectiveness of our method in improving the model performance with higher time efficiency than existing baselines. 
\end{abstract}
\renewcommand{\thefootnote}{\fnsymbol{footnote}}
\footnotetext[1]{Corresponding author}
\section{Introduction}
Different from traditional sentence-level Relation Extraction (RE), document-level relation extraction (DocRE) aims to extract the relations between multiple entity pairs within a document. The input documents of DocRE typically contain many named entities and are involved in multiple relation facts. Compared with sentence-level RE, DocRE is a more challenging task with richer interactions between the entity mentions within the document. Previous work in DocRE generally learns in a fully supervised manner, using human-annotated datasets with ground-truth labels for training and evaluation. However, human annotations for DocRE are more expensive than that of sentence-level RE due to the complexity of the task. Therefore, the expansion of DocRE datasets is costly and slow, which limits the application of DocRE. 

Distant supervision has already been used in RE to significantly augment the training data \cite{mintz2009distant,riedel2010modeling}. In sentence-level RE, distant supervision automatically annotates the sentences by aligning the mentioned entity pairs with the relations in the existing knowledge bases, assuming that all the co-appearing entity pairs in a sentence express their existing relations in the knowledge base. Despite the potential risk of noisy instances (instances with wrong labels), \newcite{yao-etal-2019-docred} introduces distant supervision into the construction of DocRED, the most widely used dataset in DocRE. The statistics of distantly supervised (DS) data and human-annotated data of DocRED are shown in Table \ref{table:docred}. According to the statistics, the size of DS data is much larger (about 20 times) than the human-annotated data in DocRED, indicating that DS data holds great potential to improve the performance of DocRE. However, due to the noisy nature of distant supervision and the overly large size of DS data, the utilization of the DS dataset is rarely discussed in the area of DocRE.

  \begin{table}[htbp]
    \centering
    \begin{tabular}{lcccc}
      \toprule
      Dataset & \#Doc.  & \#Ins. & \#Fact  &\#Ent.\\
      \midrule
      Annotated & 5k & 63k & 56k  & 132k\\
      Distant  & 101k  &  1,508k & 881k & 2,558k \\
      \bottomrule
    \end{tabular}
    \caption{The statistics of the human-annotated and distantly labeled datasets of DocRED \cite{yao-etal-2019-docred}. Doc., Ins., Fact and Ent. indicate the numbers of documents, relation instances, relation facts and entities respectively.}
    \label{table:docred}
  \end{table}

With the development of Pre-trained Language Models (PLMs), some of the recent work in DocRE proposes to utilize the DS data for pretraining PLMs and achieve considerable improvements \cite{tan-etal-2022-document,ma-etal-2023-dreeam,li-etal-2023-semi,sun-etal-2023-uncertainty}. However, existing methods typically use all of the DS data for pretraining, neglecting that the expansion of DS data is much faster and cheaper than that of human-annotated data. With the fast-growing size of DS data, utilizing all of it can make pretraining extremely expensive and lead to low time efficiency. Moreover, the wrong labeling problem of DS remains a major challenge and causes a lot of noise within the DS dataset. 

To improve the efficiency of DS data utilization as well as reduce the effects of noise from distant supervision, we propose Efficient Multi-Supervision (EMS) which includes two steps: (1) Document Informativeness Ranking (DIR) for data augmentation with informative DS documents and (2) noise-resistant training using Multi-Supervision Ranking-based Loss (MSRL). In DIR, We describe the valid information in a DS document as the reliable labels it contains, and we define a scoring criterion to rank the documents in DS data according to their informativeness. Later, we use the top informative subset of DS documents to augment the training data. In the training step, we extend the adaptive ranking loss \cite{zhou2021document} to a more robust and flexible form called MSRL to receive supervision from multiple sources. We consider three sources of supervision: distant supervision from automatically generated labels, expert supervision from a trained model and self supervision from the output of the training model. Distant supervision and expert supervision participate in determining the desired ranking of relation classes in the loss function. Self supervision is employed to dynamically adjust the fitting priority of the relation classes. We conduct experiments on the DocRED dataset to demonstrate our method's effectiveness. The results show that EMS can efficiently augment the training process of DocRE and the ablation study demonstrates that both DIR and MSRL play important roles in improving the performance. Our contributions are summarized below:

\begin{itemize}
    \item The proposed Document Informativeness Ranking (DIR) is the first attempt to retrieve the most informative documents from the DS dataset. It augments the training data with higher efficiency and greatly saves the time cost of DS data utilization. 
    \item We extend the previous ranking-based loss of DocRE as Multi-Supervision Ranking-based Loss (MSRL) which enables the model to combine multiple sources of supervision in the calculation of training loss. Compared with the original ranking-based loss, MSRL is more robust against the noise from incorrect labels and is flexible in handling supervision from multiple sources.
    \item We provide detailed experiments and efficiency analysis for EMS. The experiments and analysis show that EMS can improve the training of DocRE with high efficiency.
\end{itemize}

\section{Related Work}
Relation Extraction(RE) has been a long-discussed topic in information extraction. Traditional RE mostly extracts relations between an entity pair within a sentence \cite{zeng2014relation,wang2016relation,zhang-etal-2017-position}. However, it is shown by prior works that a large number of relation facts can only be extracted from multiple sentences \cite{verga-etal-2018-simultaneously,yao-etal-2019-docred}. Therefore, various methods have been proposed to explore document-level relation extraction (DocRE) recently. Early methods in DocRE are mostly based on Graph Neural Networks \cite{scarselli2008graph}. \newcite{quirk-poon-2017-distant} first introduces document-level graphs, in which they use words as nodes and dependency information as edges. Later graph-based methods \cite{peng2017cross,song-etal-2018-n,jia-etal-2019-document,christopoulou-etal-2019-connecting,nan-etal-2020-reasoning,zeng-etal-2021-sire} typically extends the GNN architectures to learn better representations for the entity mentions. Recently, transformer-based methods, especially those with pretrained language models, have become popular since they can automatically learn the dependency information \cite{verga-etal-2018-simultaneously,wang2019fine,tang2020hin,ye-etal-2020-coreferential}. Particularly, \newcite{zhou2021document} proposes the adaptive thresholding loss to make the classification threshold adjustable to different entity pairs. \newcite{tan-etal-2022-document} adopts knowledge distillation to utilize the large but noisy distantly supervised data. Some recent work also leverages the DS data for better performance \cite{ma-etal-2023-dreeam,li-etal-2023-semi,sun-etal-2023-uncertainty}. 

However, previous methods typically use all the DS data for pertaining, which is of low efficiency. Therefore, we seek to utilize only the most informative part of the DS data to improve the model performance with higher efficiency. Moreover, we modify the widely used adaptive thresholding loss \cite{zhou2021document} to a generalized form integrating multiple sources of supervision to mitigate the noisy instance problem in DS data.

\section{Methodology}

\begin{figure*}[htbp]
    \centering
    \includegraphics[width=0.95\textwidth]{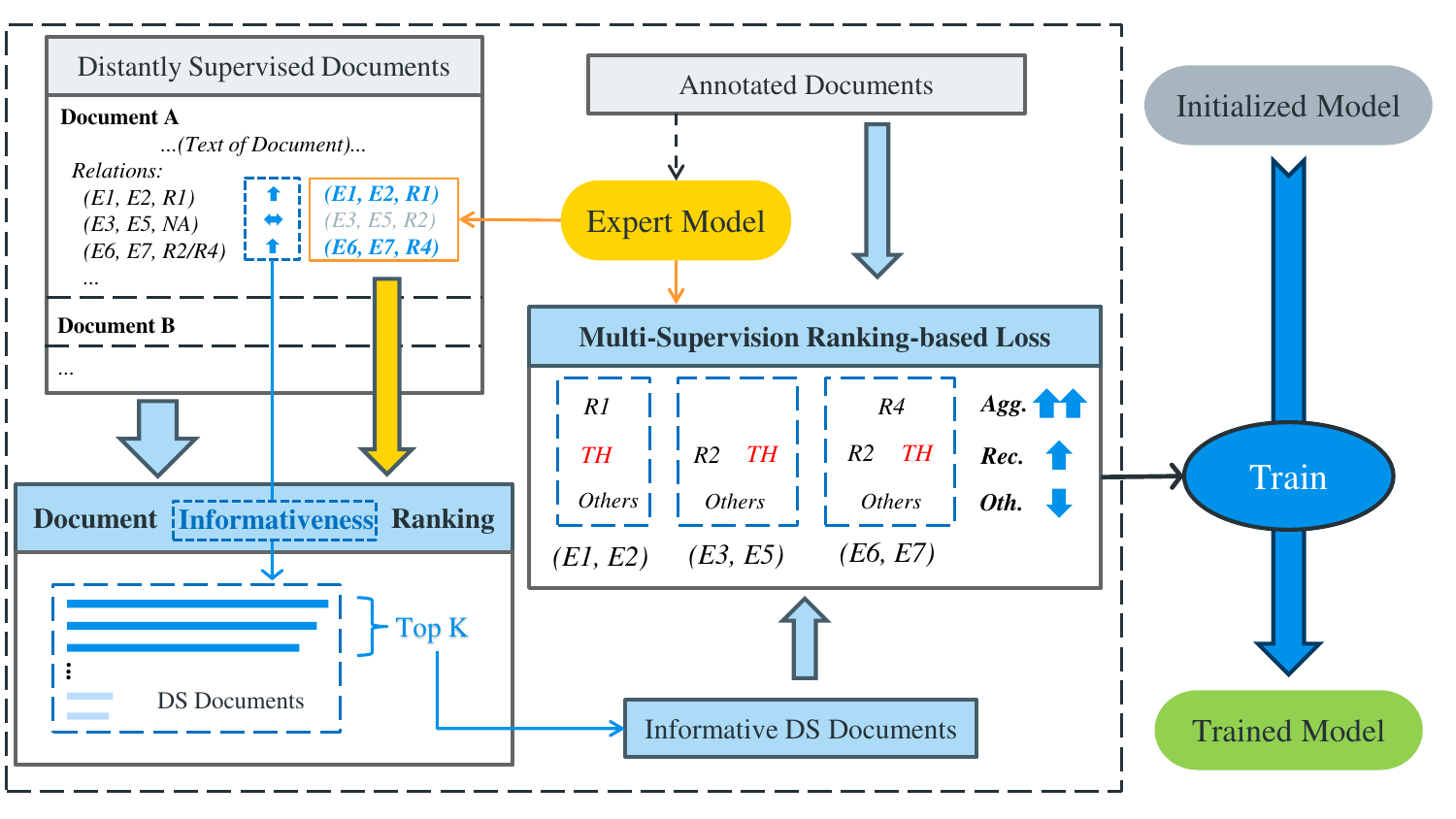}
    \caption{The illustration of EMS, contains two main components: DIR and MSRL. In MSRL, \textbf{Agg.} represents \textit{aggreements}, \textbf{Rec.} represents \textit{recommendations} and \textbf{Oth.} represents \textit{others}. }
    \label{fig.model}
\end{figure*}

The overall framework of EMS, our proposed method, is illustrated in Figure \ref{fig.model}. In both DIR and MSRL, we adopt a pretrained expert model to provide an extra source of supervision by making predictions on the DS data. First, we retrieve a set of the most informative documents from the DS dataset using Document Informativeness Ranking (DIR) to augment the training data of DocRE. Then, the model is trained using the augmented training data with the help of Multi-Supervision Ranking-based Loss (MSRL). MSRL enhances the model’s training by ensuring the relation classes adhere to specific rankings based on their logits. It also employs self supervision to dynamically adjust the learning of relation classes. 

\subsection{Preliminary}

The task of document-level relation extraction is to predict the relation classes between pairs of entities 
$(e_s, e_o)_{s, o=1...n; s \neq o}$ given a document $D$ containing the entity set $\{e_i\}_{i=1}^n$. where $e_s$ and $e_o$ represent the subject and object respectively. The set of predefined relation classes is $R \cup \{NA\}$, where $NA$ stands for \textit{no relation} between the entity pair. During the testing process, the relations between all the possible entity pairs $(e_s, e_o)_{s, o=1...n; s \neq o}$ are predicted and there may be multiple relation classes between $e_s$ and $e_o$. Each of the entity pairs is called an instance in the following parts. 

The annotation process of distantly supervised data is based on the assumption that if two entities participate in a relation, any document that contains those two entities expresses that relation \cite{mintz2009distant,yao-etal-2019-docred}. This assumption is too strong and causes the unreliability of DS labels. In order to provide an extra source of automatic annotation, we utilize a trained expert model to make predictions on the distantly supervised dataset. The relation triples provided by distant supervision are denoted as $(e_s, e_o,r_{DS})$ and the expert predictions are denoted as $(e_s, e_o,r_{EX})$, where both $r_{DS}$ and $r_{EX}$ are within $R \cup \{NA\}$ and may indicate multiple relation classes.

\subsection{Document Informativeness Ranking}

The information within distantly labeled documents is hard to obtain due to its noisy nature. Still, it holds the potential to improve the performance of DocRE models and most of the relevant methods use all the DS data for pretraining before fine-tuning on the annotated set \cite{tan-etal-2022-document,ma-etal-2023-dreeam,sun-etal-2023-uncertainty}. However, using all of the DS data is often of low efficiency. As shown in Table \ref{table:docred}, the size of existing DS data is far larger than the annotated dataset. Moreover, the automatic annotation process of DS data enables it to expand much faster and cheaper than the human-annotated dataset. Pretraining using the DS dataset before refining the model with the human-annotated dataset can help with performance gains, but this approach makes the whole process too expensive in time cost for realistic use. 

To overcome this challenge and efficiently utilize the DS dataset for the DocRE task, we propose the Document Informativeness Ranking (DIR) to retrieve the most informative subset of the DS dataset to augment the training data of DocRE. We argue that relying solely on the DS label is inevitably biased and at least one extra source of reference labels should be introduced. Therefore, we employ an expert model to automatically generate predictions on the DS data, which can be pretrained on an annotated dataset to increase efficiency. For a fair comparison in the experiments, the expert model and the training model share identical network architectures.

Based on the consistency between the DS labels and the expert predictions, we divide the relation classes of each instance into three groups: 

\begin{itemize}
    \item Agreements (Agg.): relation classes indicated by both the DS label and the expert prediction.
    \item Recommendations (Rec.): relation classes indicated by either the DS label or the expert prediction, but not both.
    \item Others (Oth.): the rest of the relation classes (not indicated by either the DS label or the expert prediction).
\end{itemize}
Distant supervision typically aligns the entities with existing pairs in the knowledge base, while the expert makes predictions based on the knowledge it has learned (from the human-annotated dataset). In other words, distant supervision is a source of prior knowledge while the expert is usually context-based. Thus, \textit{agreements} can be seen as relatively reliable labels as they are supported by both prior knowledge and contextual information. \textit{Recommendations} have discrepancies between prior knowledge and the context, which could be attributed to either the incompleteness of existing knowledge or potential biases held by the expert. In either case, the document may still express the relation classes in the \textit{Recommendations} group. However, the relation classes in the \textit{Others} group lack grounding in prior knowledge or context, making their presence in the document less probable.

The informativeness of a document can be described as the amount of \textbf{reliable} and \textbf{valuable} information it contains. In our work, each DS instance $(e_s, e_o,r_{DS})$ is considered an individual information contributor, and the amount of information it contributes is determined by the reliability and scarcity of its label. Considering the presence of two labeling sources (distant supervision and expert prediction), we propose the equation below as an attempt to quantify a document’s informativeness: 

  \begin{equation}
I(D) = \sum^{N_{D}}_{i=1}\sum^{N_r}_{j=1}V_{r_j}(y_{DS}^{ij} \cdot y_{EX}^{ij})P_{EX}^{ij}
 \end{equation} 

where $y_{DS}^{ij}$ and $y_{EX}^{ij}$ are the one-hot labels from distant supervision and expert prediction respectively. $P_{EX}^{ij}$ is the softmaxed output distribution from the expert. $N_{D}$ is the number of entity pairs in the document. $N_r$ is the number of predefined relation classes $R$, note that $NA$ is not considered in DIR. $V_{r}$ is a class-weighting vector to encourage the retrieval of the instances expressing rare relation classes. In the experiments, we directly apply the class weight function of scikit-learn \footnote{https://scikit-learn.org/} toolkit based on the distribution of classes in the human-annotated dataset. After computing informativeness, we retrieve a subset of the documents with the largest informativeness $I(D)$ in the DS dataset, forming an augmentation set $S_{aug}$ where $S_{aug} \subset S_{DS}$. During training, the augmentation set $S_{aug}$ is mixed with the human-annotated dataset $S_{ann}$. The DS documents in $S_{aug}$ are considered of relatively high quality, but they still contain many wrongly labeled instances to be addressed by MSRL in training. 

\subsection{Multi-Supervision Ranking-based Loss}
Ranking-based loss functions like the Adaptive-Thresholding Loss (ATL) \cite{zhou2021document} are widely used by previous DocRE methods. In ATL, an adaptive threshold $TH$ is introduced to separate the relation classes expressed by the instance (positive) and those unexpressed relation classes (negative). The goal of training is to push positive classes above the threshold and keep negative ones below the threshold, adhering to the $positive \rightarrow TH \rightarrow negative$ ranking order. However, the labels from distant supervision can be misleading and may cause a lot of false positive or false negative instances. To alleviate this issue, we extend ATL to the Multi-Supervision Ranking-based Loss (MSRL) which combines multiple sources of supervision to alleviate the effects of noisy instances.

Different from ATL, MSRL receives two sources of labels: distant supervision and expert prediction. As stated previously, the relation classes for each instance can be divided into \textit{agreements}, \textit{recommendations} and \textit{others}. Intuitively, we hope to push \textit{agreements} above threshold $TH$ and keep \textit{others} below $TH$. As for \textit{recommendations}, we also keep them above \textit{others} without additional ranking restrictions. The idea is to fit the \textit{recommendations} in a self-paced manner, hoping that reliable \textit{recommendations} can rise above the threshold $TH$ while unreliable ones stay below. 

Similar to ATL \cite{zhou2021document}, the logit vector is broken down into two parts to compute the probability vectors: 
\begin{gather}
    P^a_r = \frac{exp(O_r)\mathbbm{1}(r\in R_{agg.})}{\sum_{r'\in R_{agg.} \cup \{TH\}} exp(O_{r'})} \notag\\
    P^b_r = \frac{exp(O_r)\mathbbm{1}(r\in R_{rec.}\cup \{TH\})}{\sum\limits_{r'\in R_{rec.} \cup R_{oth.} \cup \{TH\}}exp(O_{r'})})
\end{gather}
where $P^a_r$ only involves \textit{aggreements} and the $TH$ class, with an indicator function $\mathbbm{1}$ filtering the relation classes except \textit{aggreements}. $P^b_r$ involves \textit{recommendations}, \textit{others} and the $TH$ class.

Since \textit{recommendations} potentially contain wrong labels reflecting incomplete knowledge or biases, we hope to carefully adjust their fitting priorities during training. Intuitively, the \textit{recommendations} confirmed by the current predictions $y$ of the training model and with higher probabilities $P^b_r$ are less likely to be noisy. Thus, we design an extra class weighting mechanism based on self supervision to mitigate noisy \textit{recommendations}. On the other hand, we hope to encourage the model to focus more on the under-fitted \textit{aggreements} to effectively learn reliable knowledge. Therefore, the class weighting mechanism within MSRL is also divided into two parts:
\begin{gather}
    w^a_r = \gamma_a + (1-y_r)(1-P^a_r) \notag\\
    w^b_r = \gamma_b + y_rP^b_r
    \label{equation:class weights}
\end{gather}
where $\gamma_a$ and $\gamma_b$ are the offsets of class weights which are based on the need for normalization. When $y_r$ is negative (equals 0) and $P^a_r$ is small, it indicates that the class belonging to \textit{agreements} is under-fitted. In this case, a larger $w^a_r$ can prompt the model towards better fitting of the \textit{agreements}. In contrast, $w^b_r$ only rewards those reliable \textit{recommendations} with positive predictions $y_r=1$ and large probability values $P^b_r$.

Finally, MSRL is defined in the following form: 
\begin{equation}
    L = -\sum_{r}{\log(w^a_rP^a_r)+\log(w^b_rP^b_r)}
\end{equation}
where the first term pushes \textit{agreements} above $TH$ and the second term keeps \textit{recommendations} and $TH$ above \textit{others}. Both distant and expert supervisions are involved in dividing the relation classes into \textit{agreements}, \textit{recommendations} and \textit{others}, while self supervision dynamically adjusts the learning priorities within the groups. In summary, the idea of multi-supervision not only allows MSRL to divide the relation classes in a more fine-grained manner but also enables flexibility in handling uncertainty. When using MSRL to train on human-annotated data, there are only expert supervision (human annotations) and self supervision available. In this case, there is no \textit{recommendations}, and MSRL is equivalent to a ranking-based loss with adaptive thresholds and class weights $w_r^a$ accelerating the learning of under-fitted positive relation classes. 

Different from knowledge distillation, which uses soft labels (logits) from the teacher model as an extra source of knowledge. EMS uses one-hot labels from the expert model in both DIR and MSRL. Intuitively, soft labels contain more information than one-hot labels. However, soft labels may not be accessible in some cases, for example, when using text-to-text language models or human annotators. Therefore, employing one-hot labels enables more flexible choices for the expert in real applications.

\section{Experiments}
In this section, we first introduce the dataset and experimental settings used in our experiments. Then, we provide our main experiment results and compare EMS with several strong baselines. Finally, we discuss the effects of DIR and MSRL through ablation study. 
 
\subsection{Datasets and Settings}

 \begin{table}[htbp]
    \centering
    \begin{tabular}{lc}
        \toprule
        Hyperparameter & Value  \\
        \midrule
        Batch size  & 4  \\
        Number of epochs & 30\\
        Number of relation classes $N_r$ & 96\\
        Class Weight Offsets $\gamma_a / \gamma_b$ & 1.0 / 0.9\\
        \bottomrule
    \end{tabular}
    \caption{The details of experimental settings.}
    \label{table.settings}
  \end{table}

We employ the DocRED \cite{yao-etal-2019-docred} dataset in our experiments. DocRED is a large-scale DocRE dataset constructed from Wikipedia and Wikidata. It is the most widely used dataset for DocRE so far and has the largest available DS dataset. The statistics of DocRED are already displayed in Table \ref{table:docred}. The human-annotated dataset is divided into train/dev/test sets, with 3053/1000/1000 documents respectively. We use the dev set for evaluation and choose the best model for testing. 

The base model of our experiment is ATLOP \cite{zhou2021document}, which is a popular benchmark in DocRED. We use the same ATLOP architecture for the expert model and the training model for fair comparisons. The encoder is initialized using \textit{bert-base-cased} checkpoint\cite{devlin2018bert}. Due to the limitation of infrastructure, the experiments are run using smaller batch size settings. Our model is optimized with AdamW \cite{loshchilov2017decoupled} using a 5e-5 learning rate for the encoder and 1e-4 for the classifier, with the first 6\% steps as warmup steps. Other details of hyperparameters are shown in Table \ref{table.settings}. 

The evaluation metrics are $F_1$ and Ign $F_1$. The Ign $F_1$ represents the $F_1$ score excluding the relation triples shared by the human-annotated training set.

\subsection{Compared Baselines}

We compare our EMS with several strong baselines, some of which also utilize DS data in their frameworks. ATLOP \cite{zhou2021document} proposes a localized context pooling layer to aggregate related context for entity pairs to get better entity representations and utilizes an adaptive thresholding loss function to replace the global threshold with an entity-pair-dependent threshold. ATLOP is also the expert model adopted in our experiments. SSAN\cite{xu2021entity} utilizes co-occurrence information between entity mentions and extends the standard self-attention mechanism with structural guidance. SIRE\cite{zeng-etal-2021-sire} employs a sentence-level encoder to extract intra-sentence relations and
a document encoder to extract inter-sentence relations respectively to represent two types of relations in different ways. DocuNet \cite{ijcai2021p551} regards the DocRE task as a semantic segmentation task, attempting to capture both local context information and global interdependency among triples. NCRL \cite{ijcai2022p0630} proposes a multi-label loss that prefers large
label margins between the $NA$ class and the predefined relation classes.  KD-DocRE \cite{tan-etal-2022-document} proposes an adaptive focal loss to alleviate the long-tailed problem and uses knowledge distillation to utilize the DS dataset. The compared methods all use the BERT \cite{devlin2018bert} encoder for fair comparisons. 

As for methods concerning DS data, we choose KD-DocRE for comparison, which uses all the DS data in pretraining. KD-DocRE also shares a similar network architecture with ATLOP, which makes it a good fit for comparison. We also present the result of ATLOP pretrained by DS data and fine-tuned by human-annotated data to compare with the performance and efficiency of EMS. 

\subsection{Main Results}

   \begin{table*}[htbp]
    \centering
    \setlength{\tabcolsep}{3.0mm}{
    \begin{tabular}{lccccc}
      \toprule
      \multirow{2}{*}{Model}  & \multicolumn{2}{c}{Dev} &\multicolumn{2}{c}{Test} & \multirow{2}{*}{Relative Time Cost} \\
      \cline{2-3} \cline{4-5} & $F_1$  & Ign $F_1$ & $F_1$  & Ign $F_1$  \\
        \midrule
        \multicolumn{5}{l}{\textit{Without distantly supervised data}} \\
        \midrule
      ATLOP* \cite{zhou2021document}  & 61.05 & 59.18 & 60.85 &58.71 & 1x\\
      SSAN\cite{xu2021entity} & 59.19&  57.03 & 58.16 & 55.84  & -  \\
      SIRE\cite{zeng-etal-2021-sire} & 61.60 & 59.82 & 62.05 & 60.18 & -\\
       DocuNet\cite{ijcai2021p551} & 61.83 & 59.86 & 61.86 & 59.93& -\\
       NCRL*\cite{ijcai2022p0630} & 61.10 & 59.22  & 60.91  &58.77& -\\
       KD-DocRE \cite{tan-etal-2022-document} & 62.03 & 60.08 & 62.08  &  60.04 & -\\
        \midrule
        \multicolumn{5}{l}{\textit{With all of distantly supervised data}} \\
        \midrule
         ATLOP with DS & 63.42 & 61.57 & 63.48 & 61.43 & 34x\\
        KD-DocRE with DS & \textbf{64.81} & \textbf{62.62} & \textbf{64.76}  &  \textbf{62.56} & 111x\\
        \midrule
        \multicolumn{5}{l}{\textit{With EMS}} \\
        \midrule
        ATLOP+EMS (3\% DS data)* & 62.39 & 60.56 & 62.05 & 59.88  & 4x\\
        ATLOP+EMS (30\% DS data)* & \underline{64.08} & \underline{62.11} &  \underline{63.89} & \underline{61.98}  & 13x \\
      \bottomrule
    \end{tabular}}
    \caption{Results of EMS and baselines on DocRED. Models marked with * are reproduced or implemented by us, others are from the papers. The relative time costs are estimated using the method in Appendix \ref{Time Efficiency}. \textbf{Bold} indicates the best results among the compared methods, the second best results are \underline{underlined}.}
    \label{table:results}
  \end{table*}

Table~\ref{table:results} shows the experimental results on the DocRED dataset. According to the results, DS data greatly improves the performance of DocRE models. With only 3\% of DS data, the performance of ATLOP+EMS almost surpasses the state-of-the-art non-DS methods. However, DS data significantly increases the time costs of the models due to its massive size. Taking ATLOP as an example, with DS data, the performance increases by 2.63 on test $F_1$ and 2.72 on test Ign $F_1$. However, the time cost dramatically increases to more than 30 times due to the use of all DS data in pretraining. KD-DocRED, the state-of-the-art method, also requires a substantial cost of time to achieve good performance. By retrieving informative instances and denoised training with MSRL, EMS can improve performance with higher efficiency than the baselines. Using only 3\% of DS data, ATLOP+EMS achieves 1.2 and 1.17 improvements on test $F_1$ and test Ign $F_1$ respectively, only increasing the time cost to 4 times. Using 30\% of DS data, ATLOP+EMS even surpasses ATLOP with DS pretraining by 0.41 test $F_1$ and 0.55 test Ign $F_1$. It also achieves a comparable performance to the state-of-the-art method with 13 times the cost of the original ATLOP, which is significantly smaller than KD-DocRE.

In practice, the size of DS data grows faster than the human-annotated dataset because DS labels are much cheaper and faster to obtain. Therefore, EMS can save even more time in the real application of distant supervision. 
 
\subsection{Ablation Study}

     \begin{table}[htbp]
    \centering
    \begin{tabular}{lcc}
      \toprule
      \multirow{2}{*}{Model}  & \multicolumn{2}{c}{Dev} \\
      \cline{2-3} & $F_1$ & Ign $F_1$   \\
      \midrule
      ATLOP+EMS & \textbf{62.39} & \textbf{60.56} \\
      - Self Sup. & 62.19 & 60.23 \\
      - Expert Sup. & 61.33 & 59.21 \\
      - Distant Sup. & 61.59 & 59.70 \\
      Rand+MSRL & 61.72 & 59.84  \\
      ATLOP & 61.05 & 59.18  \\
      \bottomrule
    \end{tabular}
    \caption{Ablation study of our method using top 3\% of the DS data. \textbf{Sup.} is the abbreviation of \textbf{supervision}.}
    \label{table:ablation}
  \end{table}

Table \ref{table:ablation} shows the results of the ablation study. We conduct this part of experiments on the dev set of DocRED using the top 3\% of the DS data. Removing self supervision means removing the class weights $w_r^a$ and $w_r^b$ defined in Equation \ref{equation:class weights}. This slightly affects the performance because class weights not only accelerate the learning of under-fitted \textit{agreements} but also reduce the effects of noisy \textit{recommendations}. If the size of the augmentation set increases, the effects of removing self supervision will be more significant because more noisy instances are introduced into the training process. 

 Since MSRL distinguishes \textit{agreements}, \textit{recommendations} and \textit{Others} based on the consistency between distant supervision and expert supervision, removing either of them essentially disables MSRL. Removing distant supervision leads to sole dependence on an expert model trained on a smaller dataset, which can lead to inaccurate predictions due to unseen patterns. Removing expert supervision, on the other hand, leaves a large number of noisy instances unaddressed. Thus, both distant supervision and expert supervision are crucial for MSRL. According to the results in the third and fourth row of Table \ref{table:ablation}, removing either distant supervision or expert supervision leads to a significant performance decline. 

Rand+MSRL is a variation that selects augmentation set randomly instead of on the basis of informativeness. Other settings are identical to ATLOP+EMS. The presented result is from the best model among five runs using five different random seeds. The performance decreases by 0.67 for $F_1$ and 0.72 for Ign $F_1$ compared with using DIR. The difference in performance demonstrates that DIR is effective in retrieving informative documents from the DS dataset. 

\begin{figure*}[htbp]
    \centering
    \includegraphics[width=0.84\textwidth]{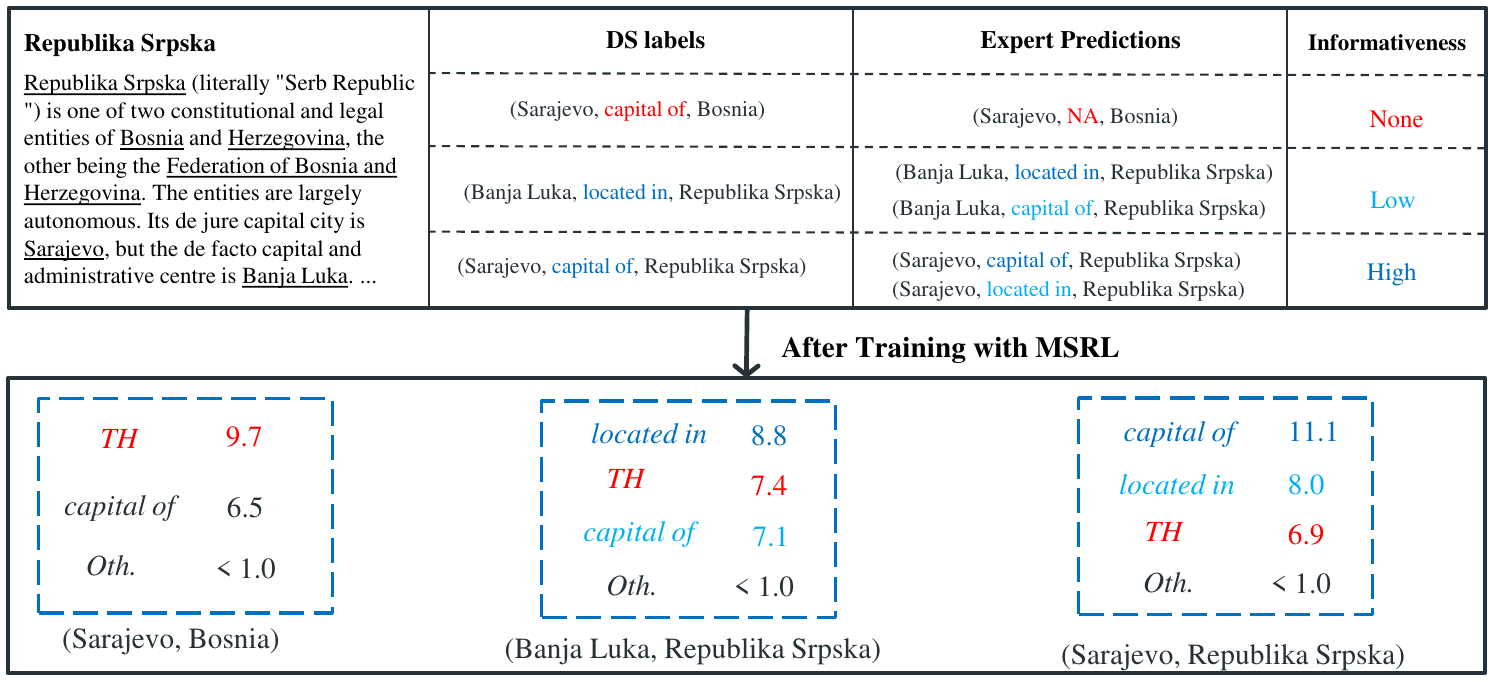}
    \caption{A retrieved document with some representative instances. The numbers are the logit values of the relation classes after training, and "located in" is the abbreviation of relation class "located in the administrative territorial entity".}
    \label{fig.case}
\end{figure*}
From the above results and discussions, we can conclude that distant supervision, expert supervision and self supervision all proved useful in the EMS framework. Also, it is clear that the two main components of EMS, DIR and MSRL, are both effective in improving the performance of DocRE.

\section{Case Study}

In order to illustrate the idea of multi-supervision, we choose an example document retrieved by DIR from the DS dataset and present it in Figure \ref{fig.case}. At the upper part of the figure, DIR estimates the informativeness of each instance. Since \textit{capital of} is a rare relation class with only dozens of instances in the human-annotated set while \textit{located in} is a more common one, the informativeness of the third instance is higher than the second one. After training on the augmented dataset $\{S_{ann} \cup S_{aug}\}$ with MSRL, the logit values of the instances are shown at the bottom part of Figure \ref{fig.case}. For the pair (Sarajevo, Bosnia), \textit{capital of} is not the gold label according to the context of the document, but distant supervision indicates that \textit{capital of} could be applicable to this entity pair in other contexts. Therefore, it is acceptable and reasonable that \textit{capital of}, which is a \textit{recommendation} from distant supervision, has a higher logit value than the classes from \textit{Others}. Regarding entity pairs (Banja Luka, Republika Srpska) and (Sarajevo, Republika Srpska), the \textit{agreements} are both far above the \text{TH} threshold. The \textit{recommendation} for (Banja Luka, Republika Srpska) is ambiguous because it is undefined whether de jure capital indicates the \textit{capital of} relation, so the logit value is near the \textit{TH} threshold. The \textit{recommendation} for (Sarajevo, Republika Srpska), \textit{located in}, is a missing gold label due to the incompleteness of distant supervision. Therefore, the logit value of \textit{located in} tends to rise above the threshold after learning from the augmented dataset. This case study illustrates the process of DIR and the outcome of MSRL and shows that integrating multiple sources of supervision enables the model to learn from DS instances with better robustness and flexibility.

\section{Conclusions} 
In this paper, we introduce EMS, an efficient and effective approach leveraging DS data to enhance DocRE models. EMS comprises two key components: DIR and MSRL. Unlike traditional methods that costly pretrain on the entire DS dataset, DIR retrieves the most informative documents from DS to create an augmentation set. Subsequently, the model undergoes training with MSRL, which flexibly mitigates noisy DS labels by integrating multiple sources of supervision. Our experiments demonstrate that EMS can significantly boost the DocRE model with higher time efficiency than existing baselines.

\section{Limitations}
Our work still has some limitations. Firstly, EMS depends on an expert model to provide an extra source of supervision, meaning that the capability of the expert is crucial to the effectiveness of EMS. Secondly, the useful information within the informative documents is still very sparse due to the highly noisy nature of distant supervision, which makes the learning on the augmentation set inefficient compared with that on the annotated set. Thirdly, though the network architecture is not likely to affect the efficacy of EMS, there is still a lack of combinations between EMS and all kinds of DocRE models in our experiments. 
\bibliography{anthology,custom}
% \bibliography{custom}

\appendix

\section{Time Efficiency}
\label{Time Efficiency}
Previous methods concerning DS data mostly involve pertraining using the whole DS data. Taking KD-DocRE\cite{tan-etal-2022-document} as an example, it first pretrains the teacher model on DS data, then inference logits for the DS data. It also needs to pretrain the student model on DS data before fine-tuning it on the human-annotated dataset. In contrast, EMS pretrains the expert on human-annotated dataset, inferences using the DS data for informativeness ranking, and trains the model with the augmented dataset. Since EMS does not need to repeatedly train on the large DS dataset, it is much more efficient in the cost of time compared with previous baselines. For better comparisons, we give a rough estimation to support our idea based on the number of processing steps. 

For convenience of estimation, we assume the processing time needed for inference or training on the same set of data is similar. Under this assumption, we further assume the time needed for one processing step in inference or training as $t$, which is the minimal unit of time in our analysis. We represent the sizes of DS data $S_{DS}$ and human-annotated data $S_{ann}$ as $M$ and $m$ respectively. With the above notations, we are able to represent the estimated time costs of DocRE methods. For example, the time cost of training the original ATLOP model for 30 epochs can be estimated as $30mt$. 

For EMS, we assume that each training round includes $\{k_n, n=1,2\}$ epochs. Then, the time cost of ATLOP+EMS can be estimated as $((k_1+k_2)m+k_2m_A+M)t$ with $m_A$ being the size of the augmentation set $S_{aug}$. By taking $\{k_n, n=1,2\}$ as $\{30,30\}$ respectively, $\frac{M}{m} \approx 33$ in DocRED, and $\frac{m_A}{m} \approx 10$ in our setting, the estimated time cost is $393mt$. We adopt the time cost of the original ATLOP ($30mt$) as the standard time cost for ease of comparison, and the relative time cost of KD-DocRE is $\frac{393mt}{30mt} \approx 13$. We estimate the relative time cost of DS-related methods using the same idea and present the results in Table \ref{table:results}.
 
Notably, ATLOP has the simplest architecture among the analyzed methods and intuitively has the shortest processing time in each training step. Therefore, the relative time costs of KD-DocRED are likely to be underestimated.

\end{document}